\documentclass{article}

\usepackage[numbers,sort&compress]{natbib}

\usepackage[preprint]{neurips_2024}
\usepackage{wrapfig}

\usepackage[utf8]{inputenc} 
\usepackage[T1]{fontenc}    
\usepackage[colorlinks]{hyperref}       
\usepackage{url}            
\usepackage{booktabs}       
\usepackage{amsfonts}       
\usepackage{nicefrac}       
\usepackage{microtype}      
\usepackage{xcolor}         
\usepackage{graphicx}
\usepackage{multirow}
\usepackage[font={small}]{caption}
\usepackage{subcaption}
\usepackage{xspace}
\usepackage{amsmath}
\usepackage{amssymb}
\usepackage{enumitem}
\usepackage{bbm}
\usepackage{pifont}
\usepackage[capitalize]{cleveref}
\let\oldding\ding
\renewcommand{\ding}[2][1]{\scalebox{#1}{\oldding{#2}}}

\usepackage{tcolorbox}
\definecolor{lightroyalblue}{HTML}{F6F8FD} 
\definecolor{royalblue}{HTML}{4169E1}
\definecolor{lighterblue}{HTML}{f2fafd}  
\newtcolorbox{abox}{colback=lightroyalblue,colframe=black}

\hypersetup{ 
    colorlinks=true,
    linkcolor=royalblue,
    filecolor=magenta,      
    citecolor=royalblue,
    pdftitle={Measuring Style Similarity in Diffusion Models},
    pdfpagemode=FullScreen,
    }

\crefname{section}{Sec.}{Secs.}
\Crefname{section}{Section}{Sections}
\Crefname{table}{Table}{Tables}
\crefname{table}{Tab.}{Tabs.}

\def\etc{\emph{etc.}}
\def\vs{\emph{vs.}}

\newcommand{\oursbig}{Contrastive Style Descriptors\xspace}
\newcommand{\oursshort}{CSD\xspace}
\newcommand{\oursdata}{LAION-Styles\xspace}

\DeclareMathOperator{\x}{\mathbf{x}}

\title{Measuring Style Similarity in Diffusion Models}

\author{Gowthami Somepalli$^\star$\textsuperscript{\rm {\color{red}\ding[1.2]{169}}}, 
Anubhav Gupta$^\star$\textsuperscript{\rm {\color{red}\ding[1.2]{169}}},
Kamal Gupta \textsuperscript{\rm {\color{red}\ding[1.2]{169}}}, Shramay Palta \textsuperscript{\rm {\color{red}\ding[1.2]{169}}},\\
\and
\textbf{Micah Goldblum} \textsuperscript{\rm {\color{black}\ding[1.2]{168}}}, 
\textbf{Jonas Geiping} \textsuperscript{\rm {\color{royalblue}\ding[1.2]{171}}}, 
\textbf{Abhinav  Shrivastava} \textsuperscript{\rm {\color{red}\ding[1.2]{169}}},
\textbf{Tom Goldstein}\textsuperscript{\rm {\color{red}\ding[1.2]{169}}}
\\
\and 
\textsuperscript{\rm {\color{black}\ding[1.2]{168}}} New York University\\
\and
\textsuperscript{\rm {\color{royalblue}\ding[1.2]{171}}} ELLIS Institute, MPI for Intelligent Systems\\
\and
\textsuperscript{\rm {\color{red}\ding[1.2]{169}}} University of Maryland, College Park
}

\begin{document}

\maketitle

\def\thefootnote{}\footnotetext{\vspace{-.1in}\\$^\star$Equal contribution. Correspondence: \url{gowthami@cs.umd.edu}.}
\def\thefootnote{\arabic{footnote}}

\vspace{-.2in}

\begin{abstract}
Generative models are now widely used by graphic designers and artists. Prior works have shown that these models remember and often replicate content from their training data during generation. Hence as their proliferation increases, it has become important to perform a database search to determine whether the properties of the image are attributable to specific training data, every time before a generated image is used for professional purposes. Existing tools for this purpose focus on retrieving images of similar {\em semantic content}.  Meanwhile, many artists are concerned with {\em style} replication in text-to-image models. We present a framework for understanding and extracting style descriptors from images. Our framework comprises a new dataset curated using the insight that style is a subjective property of an image that captures complex yet meaningful interactions of factors including but not limited to colors, textures, shapes, \etc We also propose a method to extract style descriptors that can be used to attribute style of a generated image to the images used in the training dataset of a text-to-image model. We showcase promising results in various style retrieval tasks. We also quantitatively and qualitatively analyze style attribution and matching in the Stable Diffusion model. Code and artifacts are available at \url{https://github.com/learn2phoenix/CSD}.
\end{abstract}
\begin{figure}[h]
    \centering
    \includegraphics[width=0.98\linewidth]{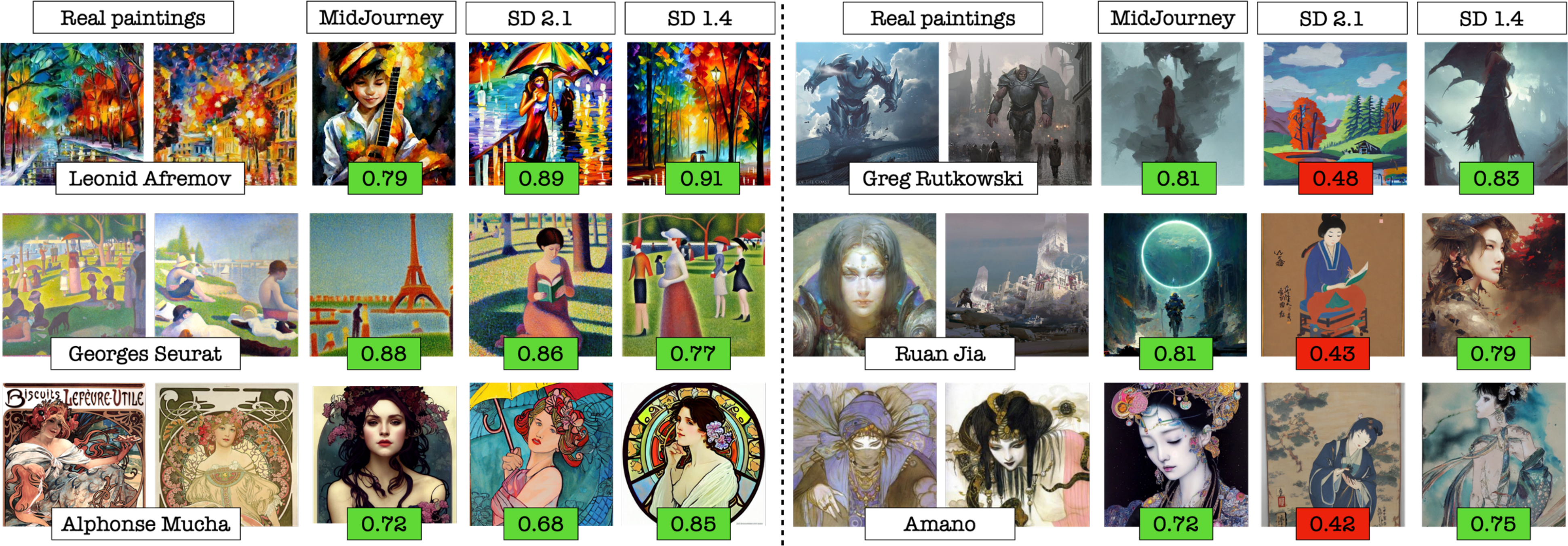}
    \caption{Original artwork of 6 popular artists and the images generated in the style of these artists by three popular text-to-image generative models. The numbers displayed below each image indicates the similarity of generated image with artist's style using proposed method. A high similarity score suggests a strong presence of the artist's style elements in the generated image. Based on our analyses, we postulate that three artists on the right were removed (or unlearned) from SD 2.1 while they were present in MidJourney and SD 1.4. Please refer to \Cref{sec:sd_96artist_analysis} for more details.} 
    \label{fig:artist_proto_analysis_qual}
\end{figure}

\section{Introduction}
Diffusion-based image generators like Stable Diffusion \cite{rombach2022high}, DALL-E \cite{ramesh2021zero} and many others \cite{betker2023improving,pernias2023wurstchen,firefly,midjourney} learn artistic styles from massive captioned image datasets that are scraped from across the web \cite{schuhmann2022laion}.
Before a generated image is used for commercial purposes, it is wise to understand its relationship to the training data, and the origins of its design elements and style attributes.
Discovering and attributing these generated images, typically done with image similarity search, is hence becoming increasingly important. Such dataset attribution serves two purposes. It enables users of generated images to understand potential conflicts, associations, and social connotations that their image may evoke. It also enables artists to assess whether and how generative models are using elements of their work.

Despite a long history of research~\cite{tenenbaum1996separating}, recovering style from an image is a challenging and open problem in Computer Vision. Many existing retrieval methods~\cite{radford2021learning,caron2021emerging,pizzi2022self} for large training datasets focus primarily on matching {\em semantic} content between a pair of images. 
Understanding the origin of the {\em style} present in a generated image, however, is much less well understood. 
To address this gap, we propose a self-supervised objective for learning style descriptors from images.  Standard augmentation-based SSL pipelines (e.g. SimCLR and variants) learn feature vectors that are invariant to a set of augmentations. Typically, these augmentations preserve semantic content and treat style as a nuisance variable.  In contrast, we choose augmentations that preserve stylistic attributes (colors, textures, or shapes) while minimizing content.  Unfortunately, SSL is not enough, as style is inherently subjective, and therefore a good style extractor should be aligned with human perceptions and definitions of style. 
For this reason, we curate a style attribution dataset, \oursdata, in which images are associated with the artist that created them. 

By training with both SSL and supervised objectives, we create a high-performance model for representing style. 
Our model, \oursshort, outperforms other large-scale pre-trained models and prior style retrieval methods on standard datasets. Using \oursshort, we examine the extent of style replication in the popular open-source text-to-image generative model Stable Diffusion \cite{rombach2022high}, and consider different factors that impact the rate of style replication. 

To summarize our contributions, we (1) propose a style attribution dataset \oursdata, associating images with their styles, (2) introduce a multi-label contrastive learning scheme to extract style descriptors from images and show the efficacy of the scheme by zero-shot evaluation on public domain datasets such as WikiArt and DomainNet (3) We perform a style attribution case study for one of the most popular text-to-image generative models, Stable Diffusion, and propose indicators of how likely an artist's style is to be replicated.

\section{Motivation}
\label{sec:sd_96artist_analysis}

\begin{wrapfigure}[22]{r}{0.5\textwidth} 
  \centering
  \includegraphics[width=0.48\textwidth]{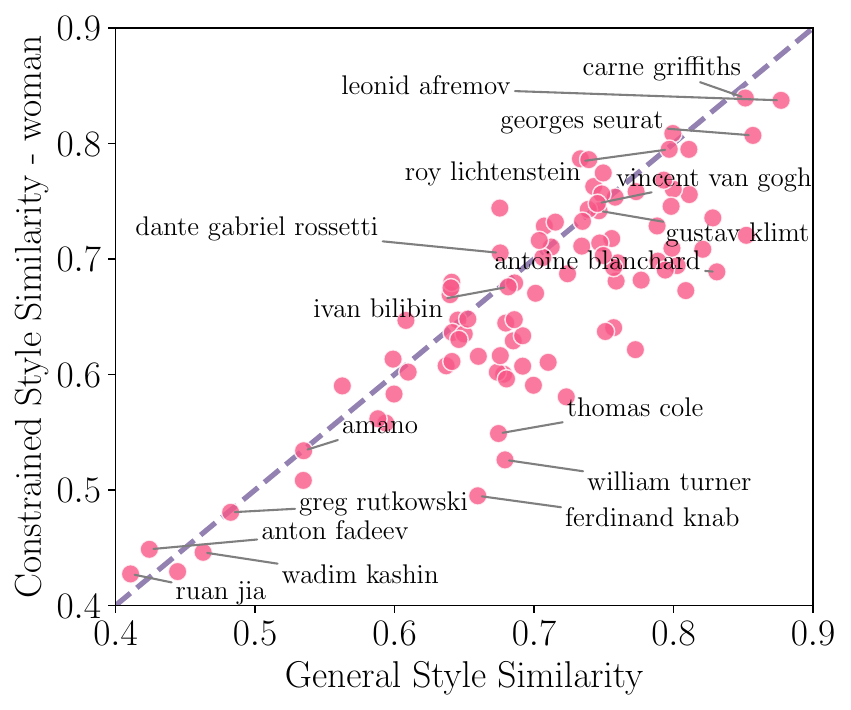}
  \caption{Style similarity of Stable Diffusion 2.1 generated images against the artist's prototypical representation. On the X-axis, the similarities are depicted when the prompt is not constrained, while the Y-axis represents similarity when the prompt is constrained to generate an image of a ``woman'' in the artist's style.}
  \label{fig:artist_clusters_scatter}
\end{wrapfigure}

We present a case study that shows how style features can be used to interrogate a generative model, and provide utility to either artists or users. We consider the task of analyzing a model's ability to emulate an artist's style, and of attributing the style of an image to an artist. We begin by curating a list of 96 artists, primarily sourced from the WikiArt database, supplemented by a few contemporary artists who are notably popular within the Stable Diffusion community\footnote{\url{https://supagruen.github.io/StableDiffusion-CheatSheet/}}. For each artist, we compute a prototype vector by averaging the embeddings of their paintings using our proposed feature extractor, \oursshort ViT-L. Next, we generate an image for each artist using Stable Diffusion 2.1 with a prompt in the format \texttt{A painting in the style of <artist\_name>}. We compute the dot product similarity between each generated image's embedding and the artist's prototype. This process was repeated multiple times for each artist, and we plot mean results in \cref{fig:artist_clusters_scatter}. We refer to this quantity as the General Style Similarity (GSS) score for an artist,  as it measures how similar a generated image is to a typical image from that artist while using our style representation model.
We also plot an analogous style similarity score, but using ``content-constrained'' prompts. For instance, one prompt template is \texttt{A painting of a woman doing <Y> style of <X>} where \texttt{X} is the name of the artist and \texttt{Y} is some setting like \texttt{reading a book} or \texttt{holding a baby} etc. See \cref{sec:study_style_wild} for all templates.  

Each point in \cref{fig:artist_clusters_scatter} represents an artist. Notice that GSS scores are highly correlated with content-constrained scores, indicating that our feature vectors represent style more than semantic content.
Our findings reveal that SD 2.1 is much more capable of emulating some artists than others. Artists like \texttt{Leonid Afremov}, \texttt{Georges Seurat} exhibit high style similarity scores, and visual inspection of generated images confirms that indeed their style is emulated by the model (\cref{fig:artist_proto_analysis_qual} - Original artwork vs SD 2.1). On the other end of the spectrum, artists such as \texttt{Ruan Jia} and \texttt{Greg Rutkowski} showed low similarity scores, and likewise the generated images bear little resemblance to the artists' work. Interestingly, after completing this study, we discovered that \texttt{Greg Rutkowski}'s work was excluded from the training data for the Stable Diffusion 2.1 model, as reported by \cite{decrypt-greg-rutkowski}. 

This demonstrates that the Style Similarity score can be used by artists to quantify how well a model emulates their style, or it can be used by users to ascertain whether a generated image contains stylistic elements associated with a particular artist. 
After a thorough inspection of the generations from 96 artists, we hypothesize that a single-image Style Similarity score below 0.5 indicates the absence of the artist's style, whereas a score above 0.8 strongly indicates its presence.

In \Cref{fig:artist_proto_analysis_qual}, we show original artworks for 6 artists, and generations from MidJourney~\cite{midjourney}, Stable Diffusion 2.1 and Stable Diffusion 1.4~\cite{rombach2022high} for each of these artists. The 3 artists on the left side have high GSS while the ones on the right side have low GSS. Below each generated image, we display the similarity against the artist's prototype vector. We see high image similarity scores in the Midjourney generations and qualitatively these images look stylistically similar to artists' original artworks. We also see the interesting cases of \texttt{Greg Rutkowski}, \texttt{Ruan Jia}, and \texttt{Amano} whose style is captured by Stable Diffusion 1.4, while being notably absent in Stable Diffusion 2.1. This finding is in line with reports suggesting that some of these artists were removed from the training data of Stable Diffusion 2.1~\cite{decrypt-greg-rutkowski}. Based on this analysis, we postulate that \texttt{Ruan Jia}, \texttt{Wadim Kashin}, \texttt{Anton Fadeev}, \texttt{Justin Gerard}, and \texttt{Amano} were also either excluded from the training data or post-hoc unlearned/removed from Stable Diffusion 2.1.

\section{What is style?}
The precise definition of ``style'' remains in contention, but many named artistic styles (e.g., cubism, impressionism, etc...) are often associated with certain artists. We leverage this social construct, and define style simply as the collection of global characteristics of an image that are identified with an artist or artistic movement. These characteristics encompass various elements such as color usage, brushstroke techniques, composition, and perspective.

\noindent\textbf{Related work.}
Early computer vision algorithms attempted to model style using low-level visual features like color histograms, texture patterns, edge detection, and shape descriptors. Other computational techniques involve rule-based systems, such as the presence of specific compositional elements, the use of specific color palettes, or the presence of certain brushstroke patterns to identify specific style characteristics~\cite{gibson1966senses, lawrence2002art, graham2012statistics,srinivasa2022wikiartvectors,matthews2020distinguishing,hughes2010stylometrics,hughes2011comparing,silva2021automatic,willats2005defining,lun2015elements,li2011rhythmic,yao2009characterizing,sablatnig1998hierarchical}.

Modern studies have focused on the task of transferring style from one image to another\cite{gatys2016image,luan2017deep,zhang2013style, wang2022fine, dumoulin2016learned, park2020swapping,huang2017arbitrary}. Some works have also concentrated on style classification~\cite{karayev2013recognizing, lu2015deep,lecoutre2017recognizing,garcia2018read,sandoval2019two,bai2021explain,chu2018image,agarwal2015genre,joshi2020art,rodriguez2018classification,menis2020deep}. A limited number of studies address in-the-wild style quantification, matching, and retrieval~\cite{matsuo2016cnn, ruta2021aladin, huang2018multimodal,wynen2018unsupervised,gairola2020unsupervised}.
In their seminal work, Gatys et al.~\cite{gatys2016image} introduced Gram Matrices as style descriptors and utilized an optimization loop to transfer style. Another approach proposed by Luan et al.~\cite{luan2017deep} includes a photorealism regularization term to prevent distortions in the reconstructed image. Zhang et al.~\cite{zhang2013style} formulated style transfer using Markov random fields. Beyond Gram-based style representation, Chu et al.~\cite{chu2018image} explored various other types of correlations and demonstrated performance variations. 

In a recent work by Lee et al~\cite{Lee_2021_CVPR}, two separate neural network modules were used -- one for image style and another for image content -- to facilitate image style retrieval. In the most recent related research, Wang et al.\cite{wang2023evaluating} developed an attribution model trained on synthetic style pairs, designed to identify stylistically similar images. In contrast to this approach, our method leverages real image pairs, curated automatically through their caption annotations. Despite our training dataset being approx. 16\% the size of training data used in \cite{wang2023evaluating}, we demonstrate that our model significantly outperforms this method on many zero-shot style matching tasks in the later sections.

\section{Creating a dataset for style attribution}
\label{sec:dataset}

While many large web datasets now exist, we need one that contains wide variations in artistic styles, and also labels that be used for downstream style retrieval evaluation. Some large-scale datasets specifically designed to handle such a challenge, like BAM~\cite{wilber2017bam}, are not available in the public domain and others like WikiArt~\cite{saleh2015large} are not large enough to train a good style feature extractor. In the following section, we propose a way to curate a large style dataset out of the LAION~\cite{schuhmann2022laion} Aesthetics 6+ dataset.

\smallskip\noindent
\textbf{LAION-Styles: A dataset for style distillation.}\label{laion-styles}
We curate our own dataset as a subset of LAION~\cite{schuhmann2022laion}. We start off with the 12M image-text pairs with predicted aesthetics scores of 6 or higher. Note that this dataset is extremely unbalanced, with some popular artists appearing much more frequently than others. Also, a large number of images are duplicated within the dataset which is a major issue for the text-to-image models trained on this data~\cite{somepalli2022diffusion,somepalli2023understanding}. Furthermore, the image captions within the data are often noisy and are often missing a lot of information. We address these challenges and propose a new subset of LAION-Aesthetics consisting of 511,921 images, and 3840 style tags, where each image can have one or more tags. We use this dataset for training our models.

We begin with a bank of styles collated in previous work for image understanding with the CLIP Interrogator~\cite{pharmapsychotic}. This bank of styles was curated based on typical user prompts for Stable Diffusion.
We combine the bank of artists, mediums, and movement references, to a shortlist of 5600 tags.
We then search for these tags in the 12M LAION-Aesthetics captions and shortlist the images that have at least one of the tags present. We further filter out the tags which have over 100,000 hits in the dataset since human inspection found that they refer to common phrases like `picture' or `photograph' that do not invoke a distinct style. After discarding images with an unavailable URL, we are left with about 1 million images and 3840 tags. We further deduplicate the images using SSCD~\cite{pizzi2022self} with a threshold of 0.8 and merge the tags of images that are near copies of each other. As a by-product, the deduplication also helps with the missing tags in the images, since we can simply merge the text labels of duplicate images. After deduplication, we are left with 511,921 images.

\section{\oursbig (\oursshort)}
\label{sec:ours}

\noindent
\textbf{Self-Supervised Learning.} Many successful approaches to SSL~\cite{walmer2022teaching} use a contrastive~\cite{chopra2005learning} approach, where two views (or augmentations) of the same image in the dataset are sampled and passed to a deep network to get their respective image descriptors.
The network is trained to maximize the similarity of two views of the same image and minimize agreement with other images in the batch.
Standard choices for augmentations include color jitter, blurring, grayscaling, \etc, as these alter the image's visual properties while preserving content. While these are good augmentations for object recognition tasks, they train the network to ignore image attributes associated with style.

Our approach relies on a training pipeline with two parts.  First, we use contrastive SSL, but with a set of augmentations that are curated to preserve style.  Second, we align our model with human perceptions of style by training on our labelled \oursdata dataset described in \Cref{sec:dataset}. 

\medskip
\noindent
\textbf{Proposed Approach.} We seek a model for extracting image descriptors that contain concise and effective style information. To be a useful, the model should be invariant to semantic content and capable of disentangling multiple styles. 

Given a dataset of $N$ labeled images $\{\x_i, l_i\}_{i=1}^{N}$, where each image can have one or more labels from a set of L labels, we define the label vector of the $i^\text{th}$ image as $l_i = (c_1, c_2, \dots, c_L)$, where each $c_k \in \{0, 1\}$. As mentioned in the previous section, our multi-label dataset consists of $N=511,921$ images and $L=3,840$ style tags. We consider a mini-batch of $B$ images.  Each of the images are passed to a Vision Transformer (ViT)~\cite{dosovitskiy2020image} backbone, and then projected to a $d-$dimensional vector. We consider two variants of ViT (ViT-B and ViT-L).   

Our style descriptors $f_\text{ViT}(\x_i) \in \mathbb{R}^d$ are then used to create a matrix of pairwise cosine similarity scores $ s_{i,j} = \cos(f_\text{ViT}(\x_i), f_\text{ViT}(\x_j))$. In order to compute our \textbf{multi-label contrastive loss} (MCL), we also compute the groundtruth similarity scores as $\hat{s}_{i,j} = \mathbbm{1}\left(l_i^\text{T}l_j\right)$, where $\mathbbm{1}$ is the indicator function that returns $1$ if any of the labels of the images ${i, j}$ match. Our final loss term reduces to:
\begin{align}
\mathcal{L}_\text{MCL} = -\hat{s}_{i,j}\log{\frac{\exp(s_{i,j})/\tau)}{\sum\limits_{k\neq j}{\exp(s_{i,k})/\tau)}}},
\end{align}
where $\tau$ is the temperature fixed during the training. 

Since our supervised dataset is modest in size, we add a self-supervised objective. We sample two ``views'' (augmentations) of each image in a batch and add a contrastive SSL term.  Standard SSL training routines (e.g., MoCo, SimCLR, BYOL \etc)  choose augmentations so that each pair of views has the same semantic content, but different style content.  These augmentations typically include Resize, Horizontal Flips, Color Jitter, Grayscale, Gaussian Blur, and Solarization~\cite{balestriero2023cookbook,geiping2022much}.  For our purposes, we depart from standard methods by excluding photometric augmentations (Gaussian Blur, Color Jitter), as they alter the style of the image.  We keep the following spatial augmentations - Horizontal Flips, Vertical Flips, Resize and Rotation as they keep style intact.

The overall loss function is a simple combination of the multilabel contrastive loss and self-supervised loss $\mathcal{L} = \mathcal{L}_\text{MCL} + \lambda \mathcal{L}_\text{SSL}$. During inference, we use the final layer embedding and the dot product to compute style similarity between any two images. In our experiments, we found that initializing weights to CLIP~\cite{radford2021learning} ViT-B and ViT-L improves performance.

\section{Results}
\label{sec:quant_results}
\noindent \textbf{Training details.} We present the results for two variants of our model \oursshort ViT-B and \oursshort ViT-L version. Both the models are initialized with respective CLIP variant checkpoints and are finetuned for 80k iterations on the \oursdata dataset on 4 A4000/A5000 GPUs. We use an SGD optimizer with momentum 0.9 and learning rate of 0.003 for the projection layer and $1e-4$ for the backbone. Our mini-batch size per GPU is 16. We use $\lambda=0.2$ and $\tau=0.1$ for the final model. The training takes about 8 hours for the base model and around 16 hours for the large model. See the Appendix for more details and ablations.

\smallskip
\noindent \textbf{Task.} 
We perform zero-shot evaluation across multiple datasets on a style-retrieval task. Following \cite{jiang2019svd,arandjelovic2016netvlad}, we split each dataset into two parts: \textit{Database} and \textit{Query}. Given a query image at test time, we evaluate whether we can find the ground-truth style in its nearest neighbors from the database. 

\smallskip
\noindent\textbf{Baselines.}
We compare our model against a recent style attribution model GDA~\cite{wang2023evaluating} which is trained via fine-tuning on paired synthetic style data, and  VGG~\cite{simonyan2014very,gatys2016image} Gram Matrices which are often used for neural style transfer applications. Further we compare with CLIP~\cite{radford2021learning} models supervised with free-form text captions, and with other self-supervised models such as DINO~\cite{caron2021emerging}, MoCo~\cite{he2020momentum}, SSCD~\cite{pizzi2022self}. We use the embeddings from the last layer for each of these models except for VGG where we use the Gram Matrix \cite{gatys2016image} of the last layer.
We skipped evaluations of \cite{ruta2021aladin,huang2018multimodal,gairola2020unsupervised} since both pre-trained models and training data are not available.

\smallskip
\noindent\textbf{Metrics.}
We do nearest neighbour searches for \textit{k~$\in[1,10,100]$}, and report Recall@k, mAP$@k$. We use the standard definitions of these metrics from the retrieval literature. Like \cite{bain2021frozen}, we define positive recall as the existence of a correct label in top-N matches and no recall when none of the top-N matches share a label with the query. Similarly, mAP is defined as average over precision at each rank in N, and then averaged over all queries.

\begin{table}[!ht]
\centering
    \caption{\textbf{mAP and Recall} metrics on DomainNet and WikiArt datasets. Our model consistently performs the best in all cases except one, against both self-supervised and style attribution baselines.}
    \label{tab:map_recall}
    \resizebox{\textwidth}{!}{
    \begin{tabular}{ l|ccc|ccc|cc|ccc} 
    \toprule
    & \multicolumn{3}{c|}{DomainNet } & \multicolumn{3}{c|}{WikiArt} & \multicolumn{2}{c|}{DomainNet } & \multicolumn{3}{c}{WikiArt} \\
    & \multicolumn{3}{c|}{(mAP$@k$) } & \multicolumn{3}{c|}{(mAP$@k$)} & \multicolumn{2}{c|}{(Recall$@k$)} & \multicolumn{3}{c}{(Recall$@k$)} \\
    \midrule
    \textbf{Method} & $1$ & $10$ & $100$ & $1$ & $10$ & $100$&  $1$ & $10$ & $1$ & $10$ & $100$ \\
    \midrule
    VGG Gram~\cite{gatys2015neural}          & -    & -    & -    & 25.9 & 19.4 & 11.4 &   -  &  -   & 25.9 & 52.7 & 80.4 \\
    DINO ViT-B/16~\cite{caron2021emerging}   & 69.4 & 68.2 & 66.2 & 44.0 & 33.4 & 18.9 & 69.4 & 93.7 & 44.0 & 69.4 & 88.1 \\
    DINO ViT-B/8~\cite{caron2021emerging}    & 72.2 & 70.9 & 69.3 & 46.9 & 35.9 & 20.4 & 72.2 & 93.8 & 46.9 & 71.0 & 88.9 \\
    SSCD RN-50~\cite{pizzi2022self}          & 67.6 & 65.9 & 62.0 & 36.0 & 26.5 & 14.8 & 67.6 & \textbf{95.0} & 36.0 & 62.1 & 85.4 \\
    MOCO ViT-B/16~\cite{he2020momentum}      & 71.9 & 71.1 & 69.6 & 44.0 & 33.2 & 18.8 & 72.0 & 94.0 & 44.0 & 69.0 & 88.0 \\
    \midrule
    CLIP ViT-B/16~\cite{radford2021learning} & 73.7 & 73.0 & 71.3 & 52.2 & 42.0 & 26.0 & 73.7 & 94.5 & 52.2 & 78.3 & 93.5 \\
    GDA CLIP ViT-B~\cite{wang2023evaluating} & 62.9	& 61.6 & 59.3 &	25.6 & 21.0	& 14.1 & 62.9 & 92.3 & 25.6 & 56.6 & 83.8 \\
    GDA DINO ViT-B~\cite{wang2023evaluating} & 69.5 & 68.1 & 66.1 & 45.5 & 34.6 & 19.7 & 69.5 &	93.4 & 45.5	& 75.8 & 89.0 \\
    GDA ViT-B~\cite{wang2023evaluating}	     & 67.1	& 65.6 & 64.2 & 42.6 & 32.2 & 18.2 & 67.1 &	93.6 & 42.6 & 67.6 & 87.1 \\
    \textbf{\oursshort ViT-B (Ours)}         & 78.3 & 77.5 & 76.0 & 56.2 & 46.1 & 28.7 & 78.3 & 94.3 & 56.2 & 80.3 & 93.6 \\
    \midrule
    CLIP ViT-L~\cite{radford2021learning}    & 74.0 & 73.5 & 72.2 &  59.4 & 48.8 & 31.5 &  74.0 & 94.8 & 59.4 & 82.9 & 95.1\\
    \textbf{\oursshort~ViT-L (Ours)}        & \textbf{78.3} & \textbf{77.8} & \textbf{76.5}  & \textbf{64.56}	& \textbf{53.82}	& \textbf{35.65} &  \textbf{78.3} & 94.5 & \textbf{64.56} & \textbf{85.73} & \textbf{95.58} \\
    \bottomrule
    \end{tabular}
    }
\end{table}

\smallskip
\noindent\textbf{Evaluation Datasets.} \emph{DomainNet}~\cite{peng2019moment} consists of an almost equal number of images from six different domains: Clipart, Infograph, Painting, Quickdraw, Real, and Sketch. Upon examination, we observed a strong stylistic resemblance between the Quickdraw and Sketch domains, leading us to exclude Quickdraw from our analysis. The dataset's content information was utilized to categorize the images into two main clusters of content classes. This clustering was achieved through the application of word2vec~\cite{mikolov2013distributed}. The images within the smaller cluster were designated as part of the Query set~(20,000 images), and images in the bigger cluster to Database~(206768 images). And the second dataset we evaluate all the models is, \emph{WikiArt}\cite{saleh2015large}. It consists of  80096 fine art images spread across 1119 artists and 27 genres. We randomly split the dataset into 64090 \textit{Database} and 16006 \textit{Query} images. We use the artist as a proxy for the style since there is large visual variation within each genre for them to be considered as independent styles. Under this setting, WikiArt is a challenging retrieval task as the chance probability of successful match is just .09\% while it is 20\% for DomainNet.

\subsection{Analyses and observations}
In \Cref{tab:map_recall}, we report the metrics for all the baselines considered and the proposed \oursshort method using $k$ nearest neighbors on 2 datasets - DomainNet and WikiArt. Note that while style and content are better separated in the case of DomainNet, WikiArt consists of more fine-grained styles and has more practical use cases for style retrieval. Loosely speaking, mAP@$k$ determines what percentage of the nearest neighbors that are correct predictions, while the recall determines what percentage of queries has a correct match in the top-$k$ neighbors. Our model \oursshort consistently outperformed all the pretrained feature extractors as well as the recent attribution method GDA~\cite{wang2023evaluating} on both WikiArt and DomainNet evaluations. Note that all models are evaluated in a zero-shot setting. We see the most gains in the WikiArt dataset which is more challenging with chance probability of only 0.09\%. When we look at mAP@1, which is same as top-1 accuracy, our base model outperforms the next best model by 5\% points on WikiArt and 4.6\% points on DomainNet. Our large model out-performs the closest large competitor by similar margins. Given the complexity of the task, these improvements are non-trivial.

We attribute the improvements to a couple of factors, (1) The multi-label style contrastive loss on our curated \oursdata dataset is quite helpful in teaching the model right styles (2) We hypothesize that these SSL models become invariant to styles because the way they were trained, but we are careful to not strip that away in our SSL loss component by carefully curating non-photometric augmentations in training. 

\begin{figure}[htbp]
\begin{minipage}[b]{0.66\linewidth}
    \centering
    \includegraphics[trim=0 0 65 0,clip,width=1\linewidth]{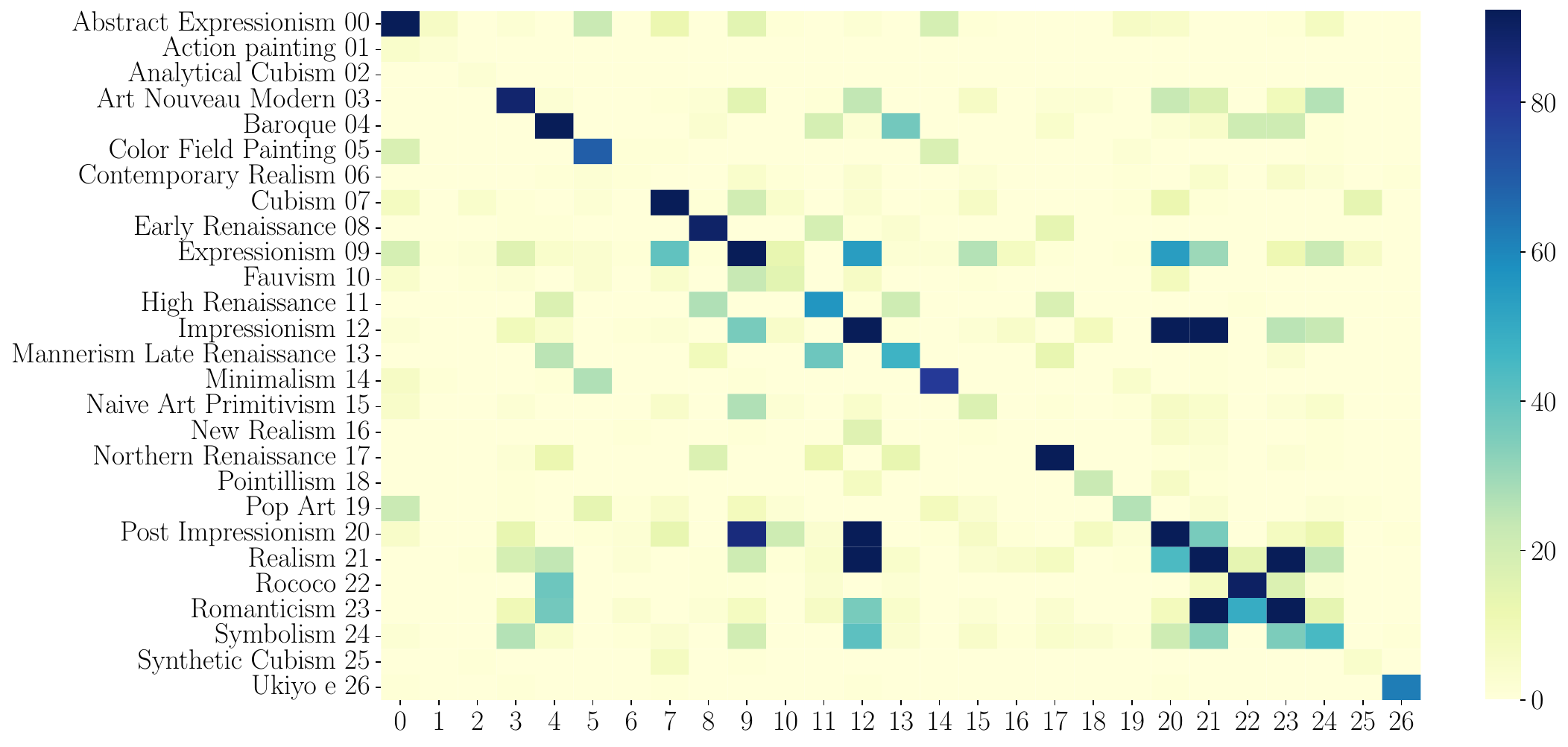}
    \caption{\textbf{Confusion Matrix of errors in WikiArt}: Art movements are predicted correctly. Errors occur in cases where movements share the same historical timeline and/or are derived from the same earlier movement.}
    \label{fig:wikiart_errors}
\end{minipage}
\hfill 
\begin{minipage}[b]{0.32\linewidth}
    \centering
    \includegraphics[width=\linewidth]{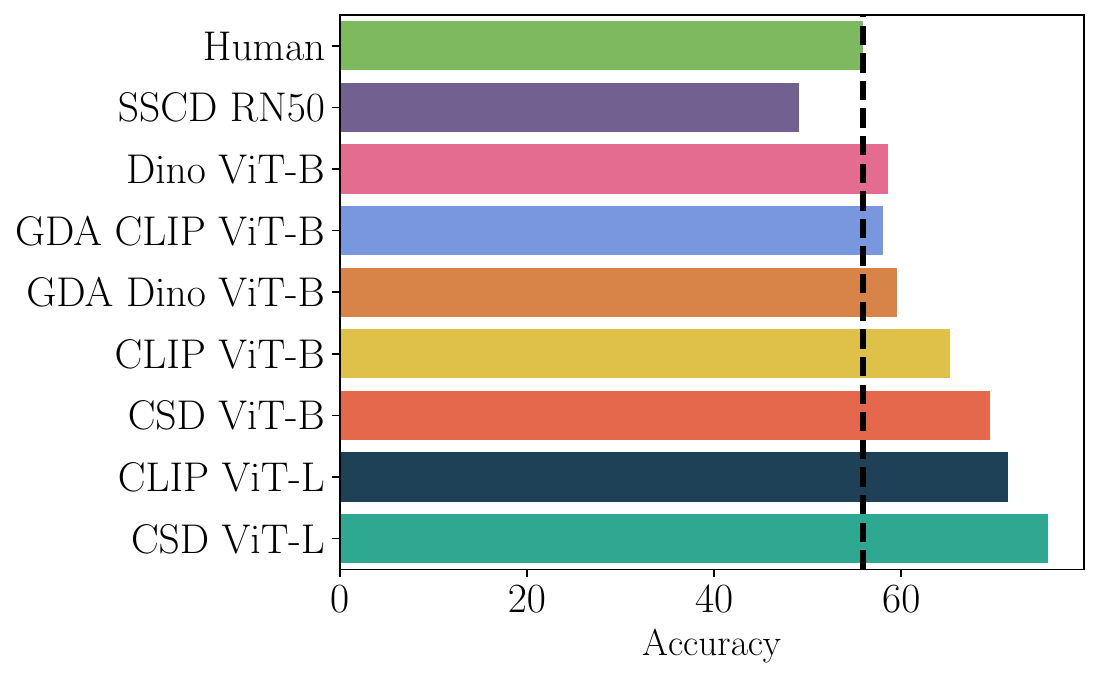}
    \caption{\textbf{Human study on Style Retrieval}: Turns out \textbf{untrained} humans are worse than many feature extractors on matching images from the same artist.}
    \label{fig:human_study}
\end{minipage}
\end{figure}

\noindent\textbf{Error Analysis.} Even though our model outperforms the previous baselines, our top-1 accuracy for the WikiArt style matching task is still at 64.56. We tried to understand if there is a pattern to these errors. For example, our model is consistently getting confused between impressionist painters \texttt{claude monet}, \texttt{gustave loiseau}, and \texttt{alfred sisley}, all of whom painted many landscapes. They depicted natural scenes, including countryside views, rivers, gardens, and coastal vistas. Another example is \texttt{pablo picasso} and  \texttt{georges braque}, who are both cubist painters. Given the impracticality of analyzing all 1,119 artists in the dataset, we opted for a macroscopic examination by categorizing errors at the art movement level. This approach is visualized in the heatmap presented in \cref{fig:wikiart_errors}. In the heatmap, we see most of the errors concentrated along the diagonal, indicating that while the model often correctly identifies the art movement, it struggles to pinpoint the exact artist.  There are instances of off-diagonal errors where the model incorrectly identifies both the artist and their art movement. For example, many Post Impressionism and Realism paintings are assigned to Impressionism artists. Upon closer examination, it becomes apparent that they closely align in terms of historical timeline and geographical origin, both being from Europe. This analysis indicates the nuanced nature of style detection and retrieval in Computer Vision. It suggests that the upper limit for accuracy in this task might be considerably lower than 100\%, even for a typical human evaluator, due to the inherent subtleties and complexities involved. 

\subsection{Human Study} 
To understand how our models compare to untrained humans, we conducted a small survey on style matching on 30 humans (excluding the authors). Following the convention in other papers~\cite{wang2022fine,karayev2013recognizing,gairola2020unsupervised,ruta2021aladin} and this paper, we assume, 2 images from same artist can be considered stylistic matches. For each query image, we gave 4 answer images out of which only one is from the same artist and hence is the right answer, so chance accuracy is 25\%. We used the Artchive dataset introduced in \cite{wang2023evaluating} to create this survey and we collected 3 responses per item to break any ties. We present the results in Fig \ref{fig:human_study}. Most interestingly, untrained humans are worse than many feature extractors at this task. SSCD is the only feature extractor that did worse than humans. Our model, \oursshort outperforms all the baselines on this dataset as well. This underpins the difficulty of style matching and also highlights the superior performance of our feature extractor.

\section{Studying style in the wild: Analysis of Stable Diffusion}
\label{sec:study_style_wild}
In the previous section, we have quantitatively shown that our model \oursbig outperforms many baselines on style matching task. Now we try to address the question, \textit{Can we do style matching on Stable Diffusion generated images?} To answer this question, we first curated multiple synthetic image collections using Stable Diffusion v 2.1~\cite{rombach2022high} and then compared them against the ``ground truth'' style matches on \textbf{LAION-Style} dataset. 
\smallskip

\begin{table}[!ht]
\vspace{-.17in}
\centering
 \caption{\textbf{mAP and Recall} of SD 2.1 generated synthetic datasets based on \textit{Simple} prompts and \textit{User-generated} prompts}
    \label{tab:sd_user_simple_map_recall}
    \resizebox{\textwidth}{!}{
    \setlength{\tabcolsep}{6pt}
\renewcommand{\arraystretch}{1.1}
    \begin{tabular}{ l|ccc|ccc|ccc|ccc} 
    \toprule
   & \multicolumn{3}{c|}{\emph{Simple} prompts} & \multicolumn{3}{c|}{\emph{User-generated} }& \multicolumn{3}{c|}{\emph{Simple} prompts} & \multicolumn{3}{c}{\emph{User-generated} } \\
   & \multicolumn{3}{c|}{(mAP$@k$) } & \multicolumn{3}{c|}{(mAP$@k$)} & \multicolumn{3}{c|}{(Recall$@k$)} & \multicolumn{3}{c}{(Recall$@k$)} \\
      \midrule
   \textbf{Method} & 1 & 10 & 100 & 1 & 10 & 100 & 1 & 10 & 100 & 1 & 10 & 100\\ 
   \midrule
      GDA - DINO & 11.6 &	10.2 &	7.6&	4.45	&4.59 &	4.24	&11.6 &	28.1	&52.83 &	4.45 &	25.22 &	67.18 \\

      CSD-ViTB & 17.53 &	16.56 &	12.68 & 5.85	& 5.96 &	5.58 & 17.53 &	38.65 &	61.85 & 5.85 &	29.26 &	74.2 \\\midrule

   CLIP ViT-L/14 & 22.3 & 20.4 & 16.1 & \textbf{6.1} & 5.7 & 5.1 & 22.3 & 44.5 & 66.2 & \textbf{6.1} & 26.0 & 71.7\\

   \oursshort (Ours) & \textbf{24.5} & \textbf{23.3} & \textbf{18.5} & 5.7 & \textbf{5.9} & \textbf{5.6} & \textbf{24.5} & \textbf{47.2} & \textbf{67.5} & 5.7 & \textbf{26.5} & \textbf{71.8}\\ 
    \bottomrule
    \end{tabular}
    }
\end{table}

\noindent\textbf{Creating synthetic style dataset.} 
The first challenge in curating synthetic images through prompts is the choice of prompts to be used for the generation. There have been no in-depth quantitative studies of the effect of prompts on generation styles. For this analysis, we chose 3 types of prompts.
\begin{enumerate}[leftmargin=0.5cm]
    \item \textit{User-generated} prompts:  We used a Stable Diffusion Prompts\footnote{\url{https://huggingface.co/datasets/Gustavosta/Stable-Diffusion-Prompts}} dataset of $80,000$ prompts filtered from Lexica.art. We used the test split and then filtered the prompts to make sure at least one of the keywords from the list we curated in Section \ref{laion-styles} is present. We then sampled 4000 prompts from this subset for query split generation.
    \item \textit{Simple} prompts: We randomly sampled 400 artists which appeared most frequently in user-generated prompts we analysed. We format the prompt as \texttt{A painting in the style of <artist-name>}, and we generate 10 images per prompt by varying the initialization seed.
    \item \textit{Content-constrained} prompts: We wanted to understand if we can detect style when we constrain the model to generate a particular subject/human in the style of an artist. For this, we used the prompt \texttt{A painting of a woman in the style of <artist-name>} or \texttt{A painting of a woman reading in the style of <artist-name>} etc., a total of 5 variations per subject repeated two times. We experimented with subjects, \texttt{woman},\texttt{dog} and \texttt{house} in this study. We provide the exact templates in the appendix.
\end{enumerate}

\begin{table}[!ht]
\centering
 \caption{\textbf{mAP and Recall} of SD 2.1 generated synthetic datasets based on \textit{Content-constrained} prompts}
    \label{tab:sd_content_ablat_map_recall}
    \resizebox{\textwidth}{!}{
    \begin{tabular}{ l|ccc|ccc|ccc|ccc|ccc|ccc} 
    \toprule
   & \multicolumn{3}{c|}{Dog} & \multicolumn{3}{c|}{House}& \multicolumn{3}{c|}{Woman} & \multicolumn{3}{c|}{Dog} & \multicolumn{3}{c|}{House}& \multicolumn{3}{c}{Woman} \\
   & \multicolumn{3}{c|}{(mAP$@k$) } & \multicolumn{3}{c|}{(mAP$@k$)} & \multicolumn{3}{c|}{(mAP$@k$)} & \multicolumn{3}{c|}{(Recall$@k$)} & \multicolumn{3}{c|}{(Recall$@k$)}  & \multicolumn{3}{c}{(Recall$@k$)} \\
   \midrule
   \textbf{Method} & 1 & 10 & 100 & 1 & 10 & 100 & 1 & 10 & 100 & 1 & 10 & 100& 1 & 10 & 100 & 1 & 10 & 100 \\ 
   \midrule
   GDA-DINO            &  2.28	& 2	&1.6&	3.9	&3.6&	2.7	&2.2&	2.3&	2&	2.28&	8.68	&28.73&	3.9&	12.3	&32	&2.2&10.1&	28.9 \\ 
   CSD-ViTB  &  4.5	& 4.31	& 3.61 &  4.6	& 4.29 & 	3.87 &  7.55 & 	7.83 & 	6.42 &  4.5	&  14.36 & 	34.88 &  4.6 & 	15.03 & 	39 & 7.55 & 	20.1 & 	42.46 \\
   \midrule
   
   CLIP ViT-L/14            & 2.3 & 2.2 & 1.9 & 4.5 & 4.2 & 3.6 & 7.4 & 7.1 & 6.2 & 2.2 & 9.8 & 29.9 & 4.5 & 13.8 & 35.3 & 7.4 & 19.0 & 41.6 \\
   
   \oursshort (Ours) & \textbf{4.9} & \textbf{4.8} & \textbf{4.2} & \textbf{6.4} & \textbf{6.2} & \textbf{5.4} & \textbf{10.8} & \textbf{10.1} & \textbf{8.6} & \textbf{4.9} & \textbf{14.5} & \textbf{34.5} & \textbf{6.4} & \textbf{17.8} & \textbf{40.6} & \textbf{10.8} & \textbf{23.4} & \textbf{44.2} \\ 
    \bottomrule
    \end{tabular}
    }
\end{table}

\begin{figure}[!t]
\centering
\includegraphics[width=1\linewidth]{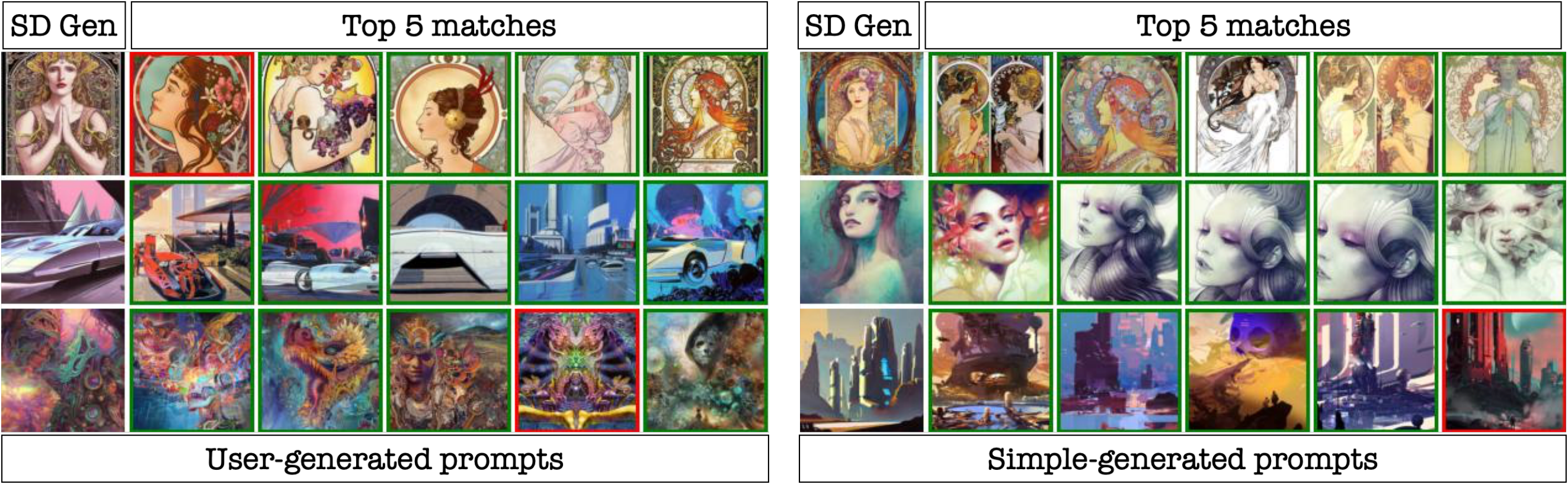}

\caption{\textbf{Nearest ``style'' neighbors.} For each generated image (referred to as SD Gen), we show the top 5 style neighbors in \oursshort using our feature extractor. The green and red box around the image indicates whether or not the artist's name used to generate the SD image was present in the caption of the nearest neighbor.}
\label{fig:sd_user_simple_csd_top10matches}
\end{figure}

We generate 4000 images for each prompt setting using Stable Diffusion 2.1. There is only one style keyword in \textit{simple} and \textit{content-constrained} prompts, which we also use as a ground truth label for matching tasks. However, \textit{user-generated} prompts can have multiple style labels within the caption, and we consider all of them as ground-truth labels.

\noindent \textbf{Style retrieval on generated images.} In \cref{tab:sd_user_simple_map_recall}, we show the retrieval results for \emph{Simple} and \emph{User-generated} prompts.
We also compare our results with the second-best performing model in the previous section, CLIP ViT-L, and a recent style attribution model GDA~\cite{wang2023evaluating}. We observe that our method outperforms CLIP on \emph{Simple} prompt dataset. For \emph{User-generated} prompts, the performance metrics are closer to CLIP model, but it's important to note that these prompts are inherently more complex. This complexity results in a different label distribution in the query set for the two types of prompts we examine, leading to varied metric ranges in each case. Additionally, our method consistently outperforms both baselines in content-constrained scenarios, as evidenced in \Cref{tab:sd_content_ablat_map_recall}. This indicates the robustness and effectiveness of our approach in dealing with a variety of prompt complexities and content specifications. We refer the reader to Appendix to understand a few caveats of this quantitative study.

\begin{figure}[!t]
\centering
\includegraphics[width=1\linewidth]{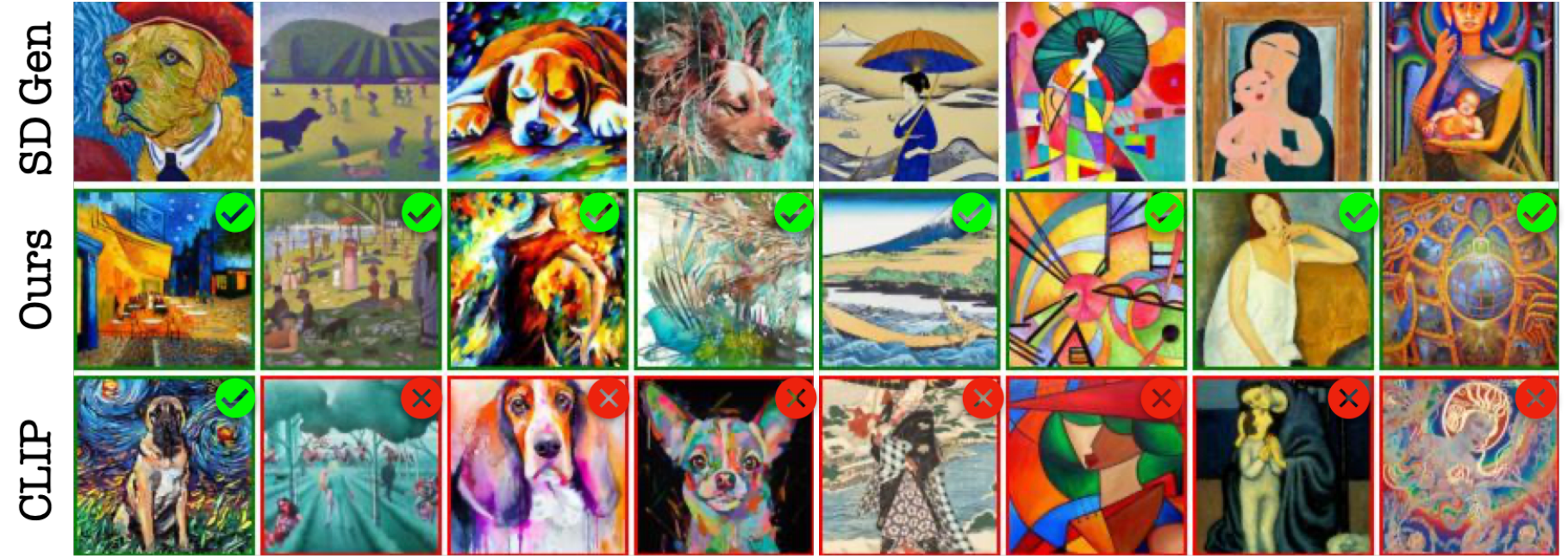}

\caption{Top row: Images generated by Stable Diffusion. Middle and Bottom rows: Top matches retrieved by CLIP vs \oursshort (ours) respectively. CLIP is consistently biased towards image content, for instance retrieving image of a dog in the Column 1, 3, 4, or image of mother and baby in Column 7 or 8. Our method emphasize less on the content but more on the image styles. Please refer to the Appendix for the prompts.
}
\label{fig:sd_clip_vs_csd_qual}
\end{figure}

\noindent \textbf{Qualitative Results.} In \cref{fig:sd_user_simple_csd_top10matches}, we showcase a selection of Stable Diffusion-generated images alongside their top 5 corresponding matches from \oursdata, as determined by the \oursshort ViT-L feature extractor. The left section of the figure displays images generated from \textit{User-generated} prompts, while the right section includes images created from \textit{Simple} prompts. To aid in visual analysis, matches that share a label with the query image are highlighted in green.  We can clearly see that the query image and the matches share multiple stylistic elements, such as color palettes and certain artistic features like motifs or textures. We observed that in generations based on user-generated prompts, perceivable style copying typically occurs only when the prompts are shorter and contain elements that are characteristic of the artist's typical content.

In Figure \ref{fig:sd_clip_vs_csd_qual}, we present several content-constrained prompt generations and their top-1 matches based on the CLIP ViT-L/14 model versus our CSD model. We observe that the CSD model accurately matches the correct artists to queries even when there is no shared content, only style. This is evident in columns 1, 3, 4, 7, and 8, where our model,\oursshort matches the correct style elements despite the subjects in the images being quite different. In contrast, the CLIP model still prioritizes content, often leading to mismatches in style.

\begin{figure}[htp]
\centering
\includegraphics[width=1\linewidth]{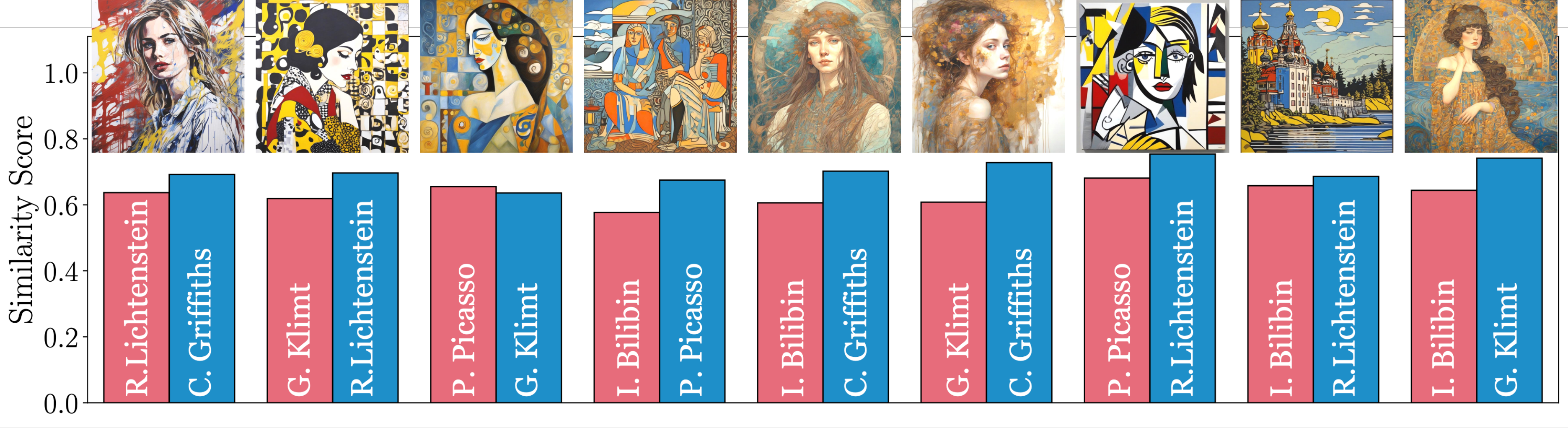}

\caption{\textbf{Does the diffusion model prefer some styles over others?} When a prompt contains two style tags, we find that SD 2.1 strongly favors the style that it can best reproduce, we suspect because of a prevalence of the style in training data. In each block, the General Style Similarity(GSS) of the left side artist (red color) is less than the right side artist (blue color). (Ref  \cref{fig:artist_clusters_scatter}). The generated image is more biased towards the artist with high GSS score.}
\label{fig:twoartist_inprompt_study}
\end{figure}

\noindent \textbf{Does the model prioritize some artists over others in the prompt?} So far in the study we primarily concentrated on the impact of including an artist's name in a style transfer prompt. In this section, we present preliminary findings in scenarios where prompts feature two artists. This is inspired by real-world user prompts from Stable Diffusion and Midjourney, where prompts often include multiple artists. We used the prompt in the style \texttt{A painting in the style of <X> and <Y>}, where \texttt{X} and \texttt{Y} represent different artists. For this study, we selected five artists with varying General Style Similarity (GSS) scores (referenced in \cref{sec:sd_96artist_analysis}). The artists, ranked by descending GSS scores, are Carne Griffiths, Roy Lichtenstein, Gustav Klimt, Pablo Picasso, and Ivan Bilibin. Note that most of the chosen artists have distinct styles that significantly differ from one another. We chose SD-XL Turbo for this analysis because it is trained on deduplicated data, reducing bias towards frequently featured artists in the training dataset.

The results for each pair of artists are presented in \cref{fig:twoartist_inprompt_study}. Interestingly, even without specific instructions to generate a female subject, most outputs depicted women, reflecting the common subject matter of the artists studied. 
We also calculated the style similarity scores for each generated image, comparing them to the prototypical styles of the artists in the prompts. In most cases, the style of the artist with the higher GSS score dominated the generated image. 
To test for potential bias towards the artist positioned first in the prompt (\texttt{X}), we conducted two trials with reversed artist positions. The results were generally consistent, with the dominant style remaining unchanged. However, in the case of Pablo Picasso and Gustav Klimt, this pattern did not hold; the model favored Picasso's cubist style over Klimt's nouveau style, possibly due to the small difference in their GSS scores. While this is not an extensive study, a consistent trend emerged: styles of artists with higher GSS scores, like Leonid Afremov and Carne Griffiths, predominantly influenced the combined style. We leave the comprehensive study on this topic to future work.

\begin{wrapfigure}[25]{r}{0.5\textwidth} 
  \centering
  \includegraphics[width=0.48\textwidth]{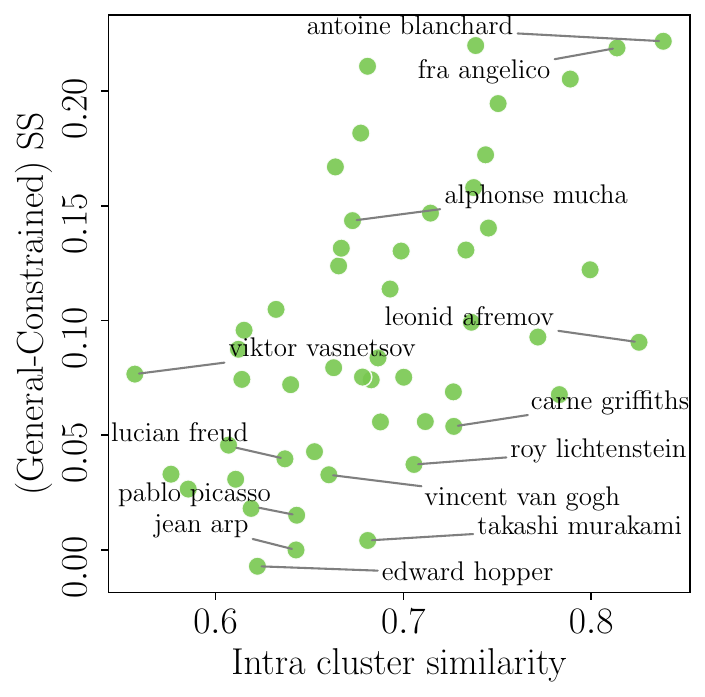}
  \caption{Style generalization to new subjects: The X-axis represents the diversity of the artists' paintings. The Y-axis shows the difference between General Style Similarity and Content-constrained Style Similarity on dog subjects.}
  \label{fig:artist_generalization}
\end{wrapfigure}
\noindent \textbf{Which styles do diffusion models most easily emulate?}
In \cref{fig:artist_clusters_scatter}, we saw that General Style Similarity(GSS) and content-constrained style similarity scores are correlated, however, we see a few artists diverging away from the identity line. How far the score is diverging away from the identity line reflects how far out-of-distribution the `subject' in the prompt is for that artist. This could serve as an indicator of the generalization capability of the artist's style. We hypothesize that artists who painted diverse subjects may have styles that generalize better to out-of-distribution (OOD) objects. To that end, we computed intra-cluster style similarity among all the artists' paintings. We plot the difference in General style similarity scores and content-constrained similarity scores and the artist-level intra-cluster similarity in \cref{fig:artist_generalization}. For this experiment, we selected `dog' as the subject, a choice informed by the observation that many artists predominantly painted women, landscapes, or cityscapes. Thus, `dog' represents a subject that is OOD, as evidenced by lower scores for the dog category compared to women or houses in \cref{tab:sd_content_ablat_map_recall}. Additionally, we limited our analysis to artists with GSS greater than 0.7 to ensure the model's proficiency in reproducing the artist's style in an unconstrained scenario. We see a high correlation of $0.568$ between these 2 variables. It seems painters of diverse subjects are more likely to have their style replicated for out of distribution objects.

In the top right corner of the figure, we note Antoine Blanchard, known for his Paris cityscapes, and Fra Angelico, who primarily focused on religious themes, including biblical scenes, saints, and other religious iconography. Conversely, in the lower left, we find Jean Arp and Pablo Picasso, whose work is characterized by abstract and non-traditional styles, encompassing a wide array of subjects. We conducted a qualitative verification to ascertain that style transfer was effective for artists located in the bottom right corner of the figure. Although this evaluation is not comprehensive, it serves as a preliminary investigation that may provide insights into the factors contributing to the generalization of style in diffusion models.

\section{Conclusion}
This study proposes a framework for learning style descriptors from both labeled and unlabeled data.  After building a bespoke dataset, LAION-Styles, we train a model that achieves state-of-the-art performance on a range of style matching tasks, including DomainNet, WikiArt, and LAION-Styles. Then, we show the substantive practical utility of this model through an investigative analysis on the extent of style copying in popular text-to-image generative models. Here, we show the model is capable of determining the factors that contribute to the frequency of style replication. Through cross-referencing of images with style copies and their original prompts, we have discover that the degree of style copying is increasing with prompt complexity. Such complex prompts lead to greater style copying compared to simple one-line prompts. This finding sheds light on the interplay between textual prompts and style transfer, suggesting that prompt design can influence the level of style copying in generative models.  Finally, note that the definition of style used in this work is strictly based on artist attribution.  We chose this definition because it can be operationalized and used in dataset construction.  This definition is certainly not a golden truth, and we look forward to future studies using alternative, or extended definitions.
\section{Acknowledgements}
This work was made possible by the ONR MURI program and the AFOSR MURI program. Commercial support was provided by Capital One Bank, the Amazon Research Award program, and Open Philanthropy. Further support was provided by the National Science Foundation (IIS-2212182), and by the NSF TRAILS Institute (2229885).

\bibliographystyle{plainnat}
\bibliography{main}

\newpage
\appendix

\centering

{\large \textbf{Measuring Style Similarity in Diffusion Models}} \\
\vspace{0.5em}{\large Appendix} \\

\raggedright

\section{Human Evaluation}
We used 30 human subjects (excluding authors) to evaluate human-level performance on the task of style matching. Participation in the study was voluntary, and none of the subjects had any prior familiarity with the task. The authors manually vetted the data presented to the subjects for the absence of any offensive or inappropriate visuals. The subjects were informed that their responses would be used to compare the human performance with ML models. We did not collect any personally identifiable information and secured an \textit{exempt} status from IRB at our institute for this study.

\section{List of artists in style analysis}
The following are the artists in the style analysis discussed in \cref{sec:sd_96artist_analysis} - \texttt{roy lichtenstein,
 justin gerard,
 amedeo modigliani,
 leonid afremov,
 ferdinand knab,
 kay nielsen,
 gustave courbet,
 thomas eakins,
 ivan shishkin,
 viktor vasnetsov,
 ivan aivazovsky,
 frederic remington,
 frederic edwin church,
 marianne north,
 salvador dali,
 pablo picasso,
 robert delaunay,
 ivan bilibin,
 rembrandt,
 frans hals,
 dante gabriel rossetti,
 max ernst,
 diego rivera,
 andy warhol,
 wadim kashin,
 caspar david friedrich,
 jan matejko,
 albert bierstadt,
 vincent van gogh,
 cy twombly,
 amano,
 anton fadeev,
 gian lorenzo bernini,
 mark rothko,
 mikhail vrubel,
 hieronymus bosch,
 katsushika hokusai,
 alphonse mucha,
 winslow homer,
 george stubbs,
 taro yamamoto,
 richard hamilton,
 carne griffiths,
 edward hopper,
 jan van eyck,
 francis picabia,
 michelangelo,
 arkhip kuindzhi,
 isaac levitan,
 gustave dore,
 antoine blanchard,
 john collier,
 paul klee,
 caravaggio,
 m.c. escher,
 leonardo da vinci,
 alan bean,
 greg rutkowski,
 jean arp,
 marcel duchamp,
 thomas cole,
 takashi murakami,
 thomas kinkade,
 raphael,
 hubert robert,
 john singer sargent,
 fra angelico,
 gustav klimt,
 ruan jia,
 harry clarke,
 william turner,
 claude monet,
 gerhard richter,
 frank stella,
 francisco goya,
 giuseppe arcimboldo,
 otto dix,
 lucian freud,
 jamie wyeth,
 rene magritte,
 titian,
 john atkinson grimshaw,
 man ray,
 albert marquet,
 mary cassatt,
 georges seurat,
 fernando botero,
 martin johnson heade,
 william blake,
 ilya repin,
 john william waterhouse,
 edmund dulac,
 peter paul rubens,
 frank auerbach,
 frida kahlo}.

\section{Model Ablations}
We conducted several ablations on the choice of the backbone for the model, the $\lambda$ hyperparameter in the loss formulation and the temperature hyperparameter, $\tau$. We present the results on the Wikiart dataset for various configurations in \cref{appendix_tab:model_ablations}. In the first ablation, we can see that CLIP ViT-B as the initialization for \oursshort gives the best performance. In the second ablation, we study the loss hyperparameter $\lambda$. $\lambda= 0$ model is trained only on Multi-label Contrastive loss, while $\lambda=\infty$ refers to model with only SSL loss. We see the best outcome when we combine both losses in the training. In the final ablation we see the effect of temperature hyperparameter. We clearly see the best outcome at $\tau=0.1$, which is conventionally the temperature used in other papers as well.

\begin{table}[h]
\centering
\captionsetup{font=small}
\caption{\textbf{Model Ablations:} We present results on Wikiart dataset. All the models are trained for same number of iterations. The baseline hyperparameters are $\lambda=0.2$ and $\tau=0.1$ and backbone initialization with CLIP ViT-B. $\bigstar$ refers to the CSD ViT-Base variant we discussed in the main paper.}
\label{appendix_tab:model_ablations}
\centering
\resizebox{0.9\textwidth}{!}{
\setlength{\tabcolsep}{8pt}
\renewcommand{\arraystretch}{1.2}
  \begin{tabular}{@{}l|l| c | c | c| c|c|c@{}}
    \toprule
     & & \multicolumn{3}{c|}{mAP@k }  & \multicolumn{3}{c}{Recall@k } \\
    \midrule
    Ablation & Variant & 1 & 10 & 100 & 1 & 10 & 100 \\
    \midrule
    \multirow{4}{10em}{Architecture,pre-training style} & SSCD RN-50& 33.2 & 24.8 & 14.11 & 33.2 & 58.8 & 83.8 \\
    &CLIP RN-50 & 51.8 & 44.2 & 25.2 & 51.8 & 77.0 & 92.1 \\
    &DINO ViT-B  & 49.8 & 39.6 & 24.4 & 49.8 & 76.3 & 92.6 \\
    &CLIP ViT-B $\bigstar$& 56.2 & 46.1 & 28.7 & 56.2 & 80.3 & 93.6 \\
   
    \midrule
    \multirow{5}{10em}{Loss Hyperparameter} & $\lambda=\infty$ & 49.1 & 40.2 & 23.9 & 49.1 & 70.3 & 85.5 \\
     & $\lambda=1$ & 52.3 & 48.1 & 26.2 &52.3 & 80.1 & 91.4 \\
     &$\lambda=0.2$ $\bigstar$  & 56.2 & 46.1 & 28.7 & 56.2 & 80.3 & 93.6 \\
     &$\lambda=0.1$ & 54.9 & 45.5 & 28.3 & 54.9 & 78.2 & 92.1  \\
     & $\lambda= 0$ & 51.8 & 44.9 & 26.6 & 51.8 & 79.5 & 90.3 \\

    \midrule
    \multirow{5}{10em}{Temperature} &$\tau=0.01$ & 55.1 & 44.4 & 27.3 & 55.1 & 79.7 & 93.6 \\
     &$\tau=0.1$ $\bigstar$& 56.2 & 46.1 & 28.7 & 56.2 & 80.3 & 93.6 \\
     &$\tau=0.5$ & 42.3 & 39.9 & 20.3 & 42.3 & 70.6 & 90.1 \\
     &$\tau=1$ & 36.2 & 28.7 & 18.0 & 36.2 & 64.4& 86.3\\

    \bottomrule
  \end{tabular}
  }
\end{table}

\section{Dataset Curation Details}


\subsection{LAION Deduplication}
During our initial retrieval experiments, we found out that most of the top-N nearest neighbors that were being returned with respect to a query image were essentially the same images. We performed deduplication of this dataset ($\sim 1.3$ million images) by computing the SSCD \cite{pizzi2022self} embedding of all the \textit{Database} images and then computing the similarity of each image with all the other images and then recursively removing the duplicate images, keeping only one image in any connected graph. The labels corresponding to the removed images were merged with the labels of the master image such that the master image has labels of all its children, the children being removed from the \textit{Database}. We considered 2 images to be the same if the inner product between their SSCD embeddings was greater than $0.8$. After the filtering, we are left with $\sim 0.5$ million images

\section{Stable Diffusion Analysis Extended}

\subsubsection{Content-constrained templates.} The following templates are used in generating content-constrained SD synthetic datasets.
\begin{itemize}
    \item \textit{Woman-constrained prompts:} (1) A painting of a woman in the style of <artist>, (2) A painting of a woman holding an umbrella in the style of <artist>, (3) A painting of a woman wearing a hat in the style of <artist>, (4) A painting of a woman holding a baby in the style of <artist>, (5) A painting of a woman reading a book in the style of <artist>
    \item \textit{Dog-constrained prompts:} (1) A painting of a dog in the style of <artist>, (2) A painting of dog playing in the field in the style of <artist>, (3) A painting of a dog sleeping in the style of <artist>, (4) A painting of two dogs in the style of <artist>, (5) A portrait of a dog in the style of <artist>
    \item \textit{House-constrained prompts:} (1) A painting of a house in the style of <artist>, (2) A painting of a house in a forest in the style of <artist>, (3) A painting of a house in a desert in the style of <artist>, (4) A painting of a house when it's raining in the style of <artist>, (5) A painting of a house on a crowded street in the style of <artist>
\end{itemize}

\subsubsection{Fig 5 prompts.}
Please find the prompts to generate the images presented in \cref{fig:sd_clip_vs_csd_qual}. The prompts are from left to right in the order.

\begin{enumerate}

    \item A painting of a dog in the style of \textbf{Van Gogh}
    \item A painting of dog playing in the field in the style of \textbf{Georges Seurat}
    \item A painting of a dog sleeping in the style of \textbf{Leonid Afremov}
    \item A painting of a dog sleeping in the style of \textbf{Carne Griffiths}
    \item A painting of a woman holding an umbrella in the style of \textbf{Katsushika Hokusai}
    \item A painting of a woman holding an umbrella in the style of \textbf{Wassily Kandinsky}
    \item A painting of a woman holding a baby in the style of \textbf{Amedeo Modigliani}
    \item A painting of a woman holding a baby in the style of \textbf{Alex Gray}
    
\end{enumerate}

\subsubsection{First observations on synthetic datasets.} When we scan through the generations, we find that for simple prompts the SD 2.1 model is borrowing the contents along with the artistic styles of an artist. For example, for the prompt \texttt{A painting in the style of Warhol} SD 2.1 always generated a version of Marilyn Diptych's painting. Similarly, all the prompts of \texttt{Gustav Klimt} generated ``The Kiss" even at different random seeds. We observed that some artist names are strongly associated with certain images and we believe this is due to dataset memorization as also discussed in \cite{somepalli2022diffusion,somepalli2023understanding}.

Another interesting issue is, for certain artists, if the prompt content diverges too vastly from their conventional ``content", the SD model completely ignores the content part sometimes and only generates the ``content" typical of the artist. For example, even when the prompt is \texttt{A painting of a woman in the style of Thomas Kinkade}, the SD model still outputs an image with charming cottages, tranquil streams, or gardens. The SD model sometimes completely ignored the content element in the prompt.

\subsubsection{Some Caveats.}
We emphasize that the above results should be taken with a grain of salt. Firstly, the LAION dataset, which we used, is inherently noisy, despite the sanitization steps we implemented as outlined in \Cref{sec:dataset}. The captions within this dataset frequently have issues. For example, tags relating to an artist or style might be absent, leading to inaccuracies in our evaluation. Even when the model correctly maps to the appropriate images in the dataset, these missing tags can cause correct results to be wrongly categorized as incorrect. Secondly, the curated style list from CLIP Interrogator is noisy. There are frequent re-occurrences of the same artist with different spellings in the style list. For example, if \texttt{Van Gogh} \vs \texttt{Vincent Van Gogh} ended up as different `style' classes, and led to a few meaningless ``errors.'' Lastly, we assume that the model strictly adheres to the prompts during generation. However, our observations indicate that in a few cases, the SD model tends to ignore the style component of a prompt and focuses predominantly on the content. This divergence results in what would have been positive matches being classified as negatives. These factors collectively suggest that while the results are informative, they should be interpreted with an understanding of the underlying limitations and potential sources of error in the data and model behavior.

\begin{figure}[h]

\subfloat[\texttt{A woman in style of <Y> 
}]{%
  \includegraphics[width=0.32\linewidth]
  {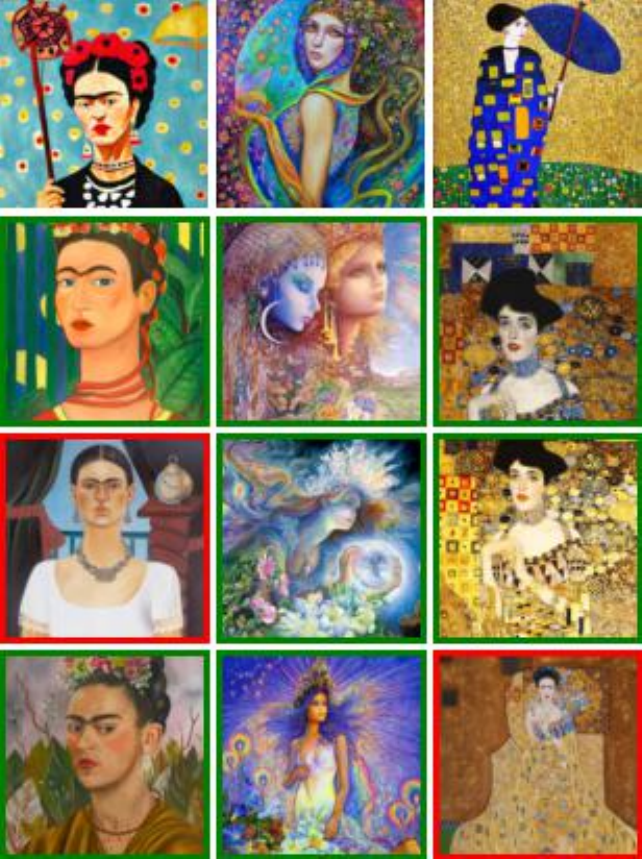}%
} \hfill
\subfloat[\texttt{A dog in style of <Y> 
}]{%
   \includegraphics[width=0.32\linewidth]{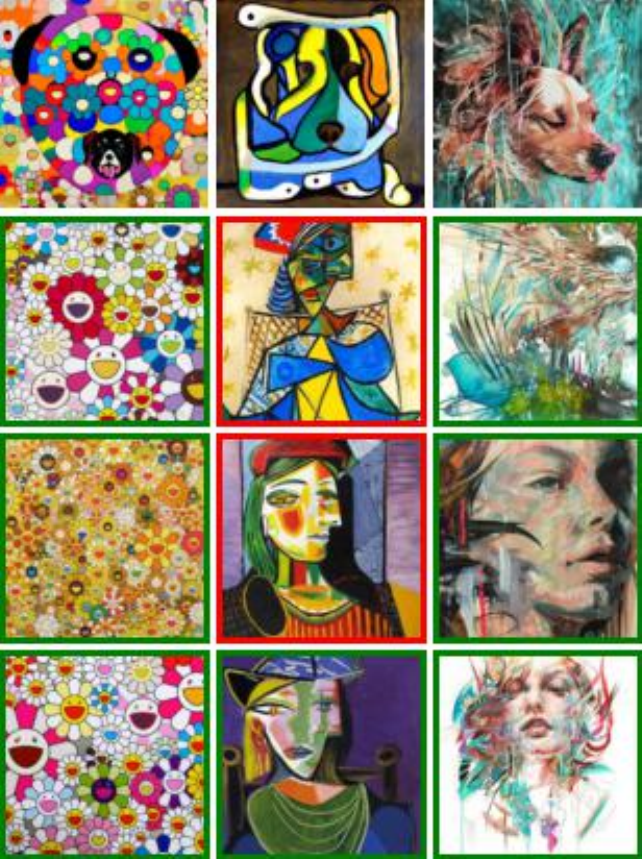}%
} \hfill
\subfloat[\texttt{A house in style of <Y> 
}]{%
   \includegraphics[width=0.32\linewidth]{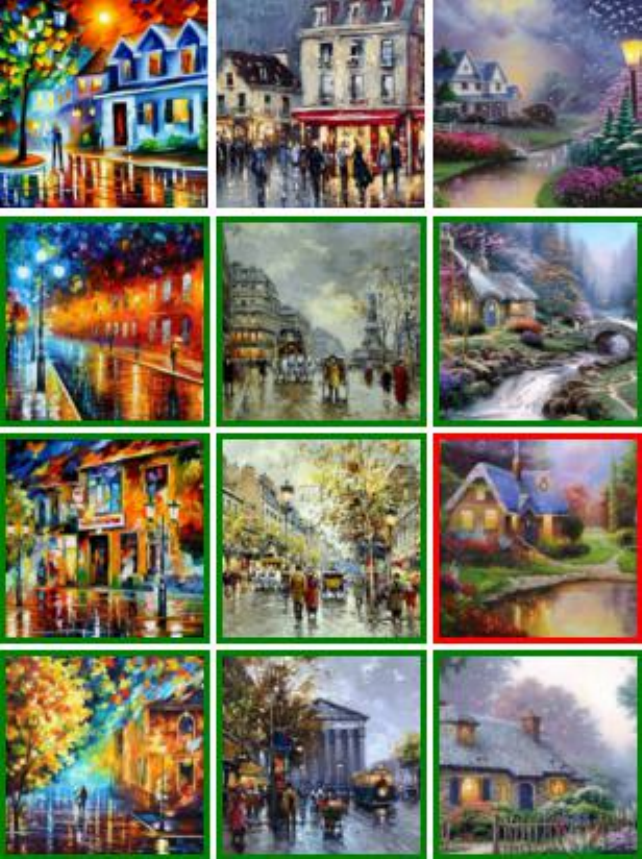}%
}

\caption{\textbf{Top row:} Images generated by Stable Diffusion~\cite{rombach2022high} using the prompt - \texttt{A <X> in the style of <Y>} where \texttt{X} comes from the set $\{\texttt{woman}, \texttt{dog}, \texttt{house}\}$ and \texttt{Y} in order are \texttt{Frida Kahlo, Josephine Wall, Gustav Klimt, Takashi Murakami, Picasso, Carne Griffiths, Leonid Afremov, Antoine Blanchard, Thomas Kinkade}. \textbf{Next two rows:} top three \emph{style} neighbors of the generated images from the LAION aesthetics datasets~\cite{schuhmann2022laion} as predicted by our model. The green and red box around the image indicate whether it was a true or false positive prediction based on whether the caption of the LAION image contained the name of the artist Y (used to generate the images).}
\label{appendix_fig:sd_gens_matches_all3}
\end{figure}

\subsubsection{Qualitative results extended.} We present a few content-constrained generations and their respective top-2 matches using CSD ViT-L feature extractor in \cref{appendix_fig:sd_gens_matches_all3}. We provide extended versions of \cref{fig:sd_user_simple_csd_top10matches} in \cref{appendix_fig:simple_caps_extended} and \cref{appendix_fig:user_caps_extended}. These figures depict the results obtained from Stable Diffusion (SD) generations using both ``simple" artist prompts and ``user-generated" prompts, along with their respective top-10 matches. The first column corresponds to the SD generation, while the subsequent columns display the identified matches.

To aid in the interpretation of these matches, we employed color-coded boxes to indicate the accuracy of the match. Specifically, green boxes represent true-positive matches, while red boxes indicate false-positive matches. However, it is important to note that the ground-truth labels assigned to the matched images may occasionally be incorrect because the ground-truth labels are generated from the LAION caption which may not always contain the artist's name. Our analysis reveals several instances of such mislabeling, particularly evident in \cref{appendix_fig:user_caps_extended}. Notably, numerous images that bear striking stylistic resemblance to the generated images are erroneously labeled as false positives.

These findings underscore the challenges involved in accurately assessing style copying in SD and emphasize the need for further exploration and refinement of evaluation methods.

\begin{figure}[t]
    \centering
    \includegraphics[width=\linewidth]{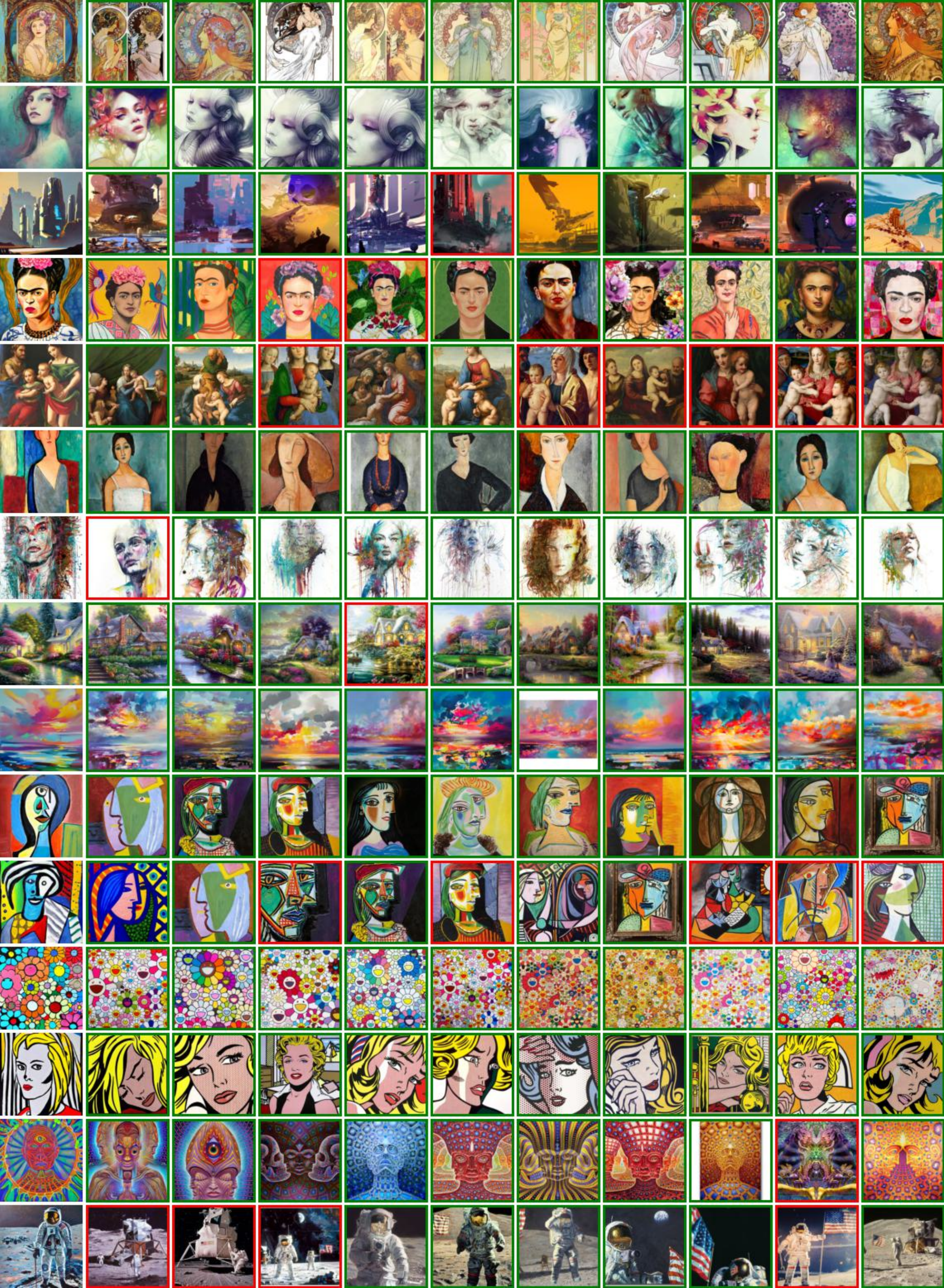}
    \caption{ \textbf{Simple caption generations and matches:} First column is SD generation, and the rest of the columns are top-10 matches in the LAION-Style database. The green box represents the correct match while the red box represents the incorrect match.} 
    \label{appendix_fig:simple_caps_extended}
\end{figure}

\begin{figure}[t]
    \centering
    \includegraphics[width=\linewidth]{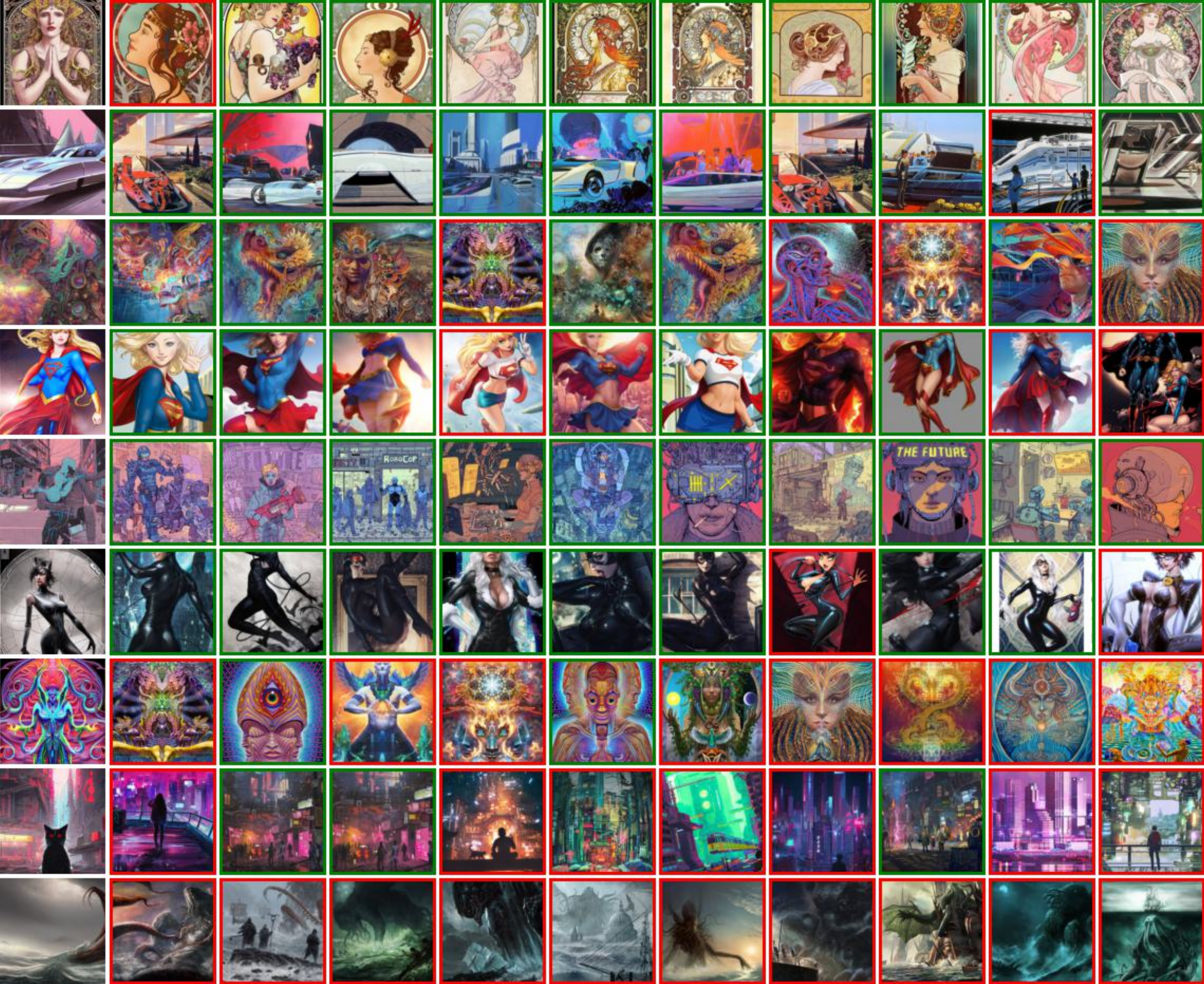}
    \caption{\textbf{\textit{User-generated} caption generations and matches:} First column is SD generation, and the rest of the columns are top-10 matches in the LAION-Style database. The green box represents the correct match while the red box represents the incorrect match. } 
    \label{appendix_fig:user_caps_extended}
\end{figure}

\end{document}